# FourCastNeXt: Optimizing FourCastNet Training for Limited Compute

**Edison Guo**[1]    **Maruf Ahmed**[1]    **Yue Sun**[1]    **Rui Yang**[1]

**Harrison Cook**[2]    **Tennessee Leeuwenburg**[2]    **Ben Evans**[1]

[1] NCI Australia, Australian National University   [2] Bureau of Meteorology, Australia

Correspondence to Edison Guo, edison.guo@anu.edu.au

**16 March 2024**

## Abstract

FourCastNeXt is an optimization of FourCastNet - a global machine learning weather forecasting model – that performs with a comparable level of accuracy and can be trained using around 5% of the original FourCastNet computational requirements. This technical report presents strategies for model optimization that maintain similar performance as measured by the root-mean-square error (RMSE) of the modelled variables. By providing a model with very low comparative training costs, FourCastNeXt makes Neural Earth System Modelling much more accessible to researchers looking to conduct training experiments and ablation studies. FourCastNeXt training and inference code are available at https://github.com/nci/FourCastNeXt.

## 1. Introduction

Recently, the Neural Earth System Model (NESM) called FourCastNet (Pathak et al., 2022b), trained on the ECMWF Reanalysis v5 (ERA5) data set, has shown impressive results on predicting various atmospheric variables globally at 0.25 degrees resolution. See Hersbach et al. (2020) for a description of the ERA5 reanalysis data set. While FourCastNet enjoys quasi-linear time and memory complexity in sequence length compared to quadratic complexity in vanilla transformers, training FourCastNet on ERA5 from scratch still requires a large amount of compute resources. FourCastNet baseline training utilized 64 NVIDIA A100 GPUs for about 16 hours (Pathak et al., 2022b, p. 7), which can be prohibitively expensive or even inaccessible to most researchers. In this work, we show methods that can train our modified version of FourCastNet using 4 NVIDIA V100 GPUs for around 35 hours, which we estimate to be around 5% of the compute required by the baseline, while maintaining comparable model performance as measured by RMSE across all 20 variables modelled (see Section 4.2 regarding computational requirements and Section 4.3.2 which lists the modelled variables). We have called our model FourCastNeXt, in a similar spirit to ConvNeXt (Liu et al., 2022).

While a thorough assessment of physical realism beyond aggregate statistics is not within the scope of this report, some discussion of physical realism is included in Section 6. Figure 1 provides an illustrative example showing that FourCastNet and FourCastNeXt at 24-hour lead-time have similar predictive skill.

In this report, we provide technical details of our methodologies along with experimental results and further analysis of different components of our methods in terms of both computational performance and physical fidelity. The methods we present are as follows:

- Increasing the size of the training set using data augmentation and alternative sampling
- Deep-norm initialization, following Wang et al. (2022)
- Using a smaller embedding patch size to better capture fine-scale image structure
- Learning the temporal flow field to improve learning efficiency
- Multi-step fine-tuning to improve model stability

This work used the resources and services of the Gadi supercomputer (NCI Australia, 2021) and Australian Research Environment (ARE) (NCI Australia, 2023b) from the National Computational Infrastructure (NCI). All the code and data related to the project, including the European Centre for Medium Range Weather Forecasts (ECMWF) ERA5 replicated data set (NCI Australia, 2020), and models (NCI Australia, 2023d), and FourCastNeXt predictions and checkpoints (NCI Australia, 2023a) are freely available.

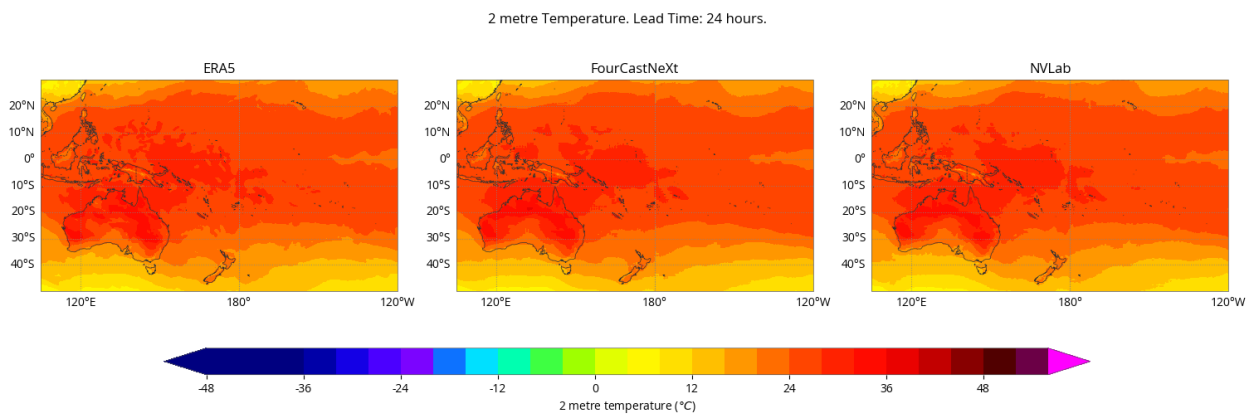

**Figure 1:** Illustrative comparison of ERA5, FourCastNeXt and NVLab's FourCastNet, in the Australian region. This shows a lead time of 24 hours, initialized at 2018-01-01T00:00. Created using `earthkit.maps` (ECMWF, 2023).

## 2. Related Work

Recent years have seen the employment and extension of a variety of deep learning models to tackle climate and weather problems, such as FourCastNet (Pathak et al., 2022b), ClimaX (Nguyen et al., 2023), Pangu-Weather (Bi et al., 2023), and GraphCast (Lam et al., 2023) - to name just a few. Each of these models is a neural network capable of performing global atmospheric modelling for weather and/or climate modelling applications. Despite their impressive performance, the computational resources required for training these models are prohibitive for many researchers and organisations.

Motivated to build efficient models under constrained resources, the deep learning community has developed a large body of literature over the past years. Notably, MobileNet (Howard et al., 2017) was the initial attempt for efficient design of convolutional networks to target mobile devices with limited CPU and memory while maintaining a similar level of performance as the baseline models. EfficientNet (Tan & Le, 2019) was developed to provide a more efficient design of ResNet (He et al., 2015) to achieve higher performance with fewer floating-point operations (FLOPS) being required.

The resource-intensive nature of neural network training has motivated the exploration of more efficient network architectures and more efficient training methods. Examples include EfficientFormer (Li et al., 2022) which improves the original Vision Transformer (Dosovitskiy et al., 2021), and ConvNeXt (Liu et al., 2022) which is based on ResNet.

While we did not apply these exact innovations, we were inspired by such results to investigate a set of methods to improve the efficiency of transformer networks as applied to the domain of climate and weather modelling.

## 3. Methods

In this section, we discuss our main modifications to FourCastNet. These enable FourCastNeXt to reach a similar level of skill to FourCastNet. Later, in Sections 4 and 5, we show that the FourCastNeXt training process converges much faster than the NVLab baseline[1] FourCastNet, allowing the overall process to use a small fraction of the original compute resources.

### 3.1 Large training set

Previous bodies of work (Hoffmann et al., 2022; Touvron et al., 2023) have shown that a large training set can help models generalize better to unseen examples.

The NVLab baseline FourCastNet was trained on a fixed set of approximately 50,000 training examples sampled from the raw ERA 5 data set. This amount of training samples, however, is only a small fraction of the raw ERA 5 data set. We follow the experimental protocols of the NVLab baseline FourCastNet to construct the training and test sets, but augment the number of samples as described below, as well as drawing on additional earlier years of data. The NVLab baseline FourCastNet was trained on the years 1979 to 2015 (both included), with validation between 2016 and 2017 and out-of-sample testing taking from 2018. Our training set spans 1959 to 2017 (both included). Our test set is taken from 2018.

Instead of having a fixed training set, we expand the training set by employing the following data augmentation strategies to construct the training examples on the fly. In other words, we do not pre-compute the training set and store them on disk but instead generate the training examples on the fly by reading the raw ERA 5 files. Our experiments utilized around 160,000 such samples during training, with the training procedure presented in Section 4. Our data augmentation approach comprised:

> **1) Randomly picking time steps**
>
> The time steps are evenly spaced for an interval of 6 hours based on standard model issuance times of midnight, 6am, midday and 6pm (UTC). For each training example, we randomly pick a time step from those times, rather than training in sequential order.
>
> **2) Random cropping over space**
>
> ERA 5 variables have a spatial size of 721 x 1440. For each training sample, we randomly crop to a region with a size of 640 x 1280.

---

[1] At the time of writing, NVIDIA provided the published model and weights under a GitHub "organization" named "NVlabs", since renamed "NVIDIA Research Projects". This can be found at https://github.com/NVlabs. We have referred to the version of the FourCastNet model published in this fashion as the NVLab baseline.

A theoretical estimation suggests that the above data augmentation strategies could produce up to (2017-1959+1) x 4 x 365 x (721-640+1) x (1440-1280+1) = 1,137,220,280 training examples. As using over 1 billion samples would have been computationally prohibitive (negating the point of this work), our experiment used this approach to generate 160,000 samples. This is more than would otherwise have been possible using only raw data, and is around three times larger than the NVLab baseline FourCastNet training set.

We also noticed that the filesystem input/output (IO) (both bandwidth and filesystem metadata latency) on the single Graphics Processing Unit (GPU) node utilized in our experiments was not fast enough to keep up with the four on-board GPUs computing speed when constructing the training examples on the fly. To address this issue and improve GPU utilization, we implemented our data loader on a Ray cluster that scales the data workers out of the GPU node.

### 3.2 Deep-norm initialization

The NVLab baseline FourCastNet used pre-normalization (pre-norm) (see e.g. Xiong et al., 2020) for its Adaptive Fourier Neural Operators (AFNO) blocks. Pre-norm is known to stabilize the early training but may result in worse performance than post-normalization (post-norm). Fortunately, recent work such as deep-norm initialization (Wang et al., 2022) has developed methods that can stabilize the training for very deep post-norm transformers, while retaining the performance of post-norm transformers. To take advantage of deep-norm initialization, we modify FourCastNet to use post-norm for the residual branches of its AFNO blocks and apply deep-norm initialization to the model weights. The results of further analysis are presented in Section 5, showing that deep-norm initialization indeed helps stabilize early training.

### 3.3 Smaller embedding patch size

The NVLab baseline FourCastNet used an 8x8 patch size for input embedding. Typical vision transformers when applied to image classification problems, for instance ImageNet (Deng et al., 2009), often only predict a single scalar (category) for an entire input image. As such, in the image classification context, an 8x8 or even 16x16 patch size can give good results (Dosovitskiy et al., 2021).

In the context of atmospheric modelling, we suggest that using a smaller patch size such as 4x4 (see e.g. Liu et al., 2021) may give superior performance. The predictions (outputs) generated by FourCastNet are entire images, matching the resolution of the input images, and thus require fine-scale structure to be well represented. We consider that a smaller patch size may be more effective in capturing this fine-scale structure, and our further analysis in Section 5 shows that a 4x4 patch size contributes to training convergence and accuracy.

Readers familiar with traditional numerical weather prediction (NWP) designs may wonder what is done in the vertical dimension and whether this forms part of the patch. While a volumetric patch may be conceivable, the various vertical levels are treated instead as separate variables and added as channels rather than treating the vertical dimension spatially. Each patch has 20 channels, one for each modelled variable. Figure 2a of the original FourCastNet paper (Pathak et al., 2022b) describes this in detail.

### 3.4 Learning the temporal flow field

The NVLab baseline FourCastNet directly predicts the outcome of the next time step (6 hours ahead) from the current time step as input. This implies the model must reconstruct the full state of the atmosphere between steps and then also apply a temporal difference. This full reconstruction may not be easily learned by the neural network, particularly in the early phase of training. Often, the temporal dynamics only result in slight changes between the neighbouring time steps. The slight changes, for example, might be a shift of a few pixels in the wind velocity fields between the neighbouring time steps. If we can capture those slight temporal deformations as much as possible, the main model will only be responsible for learning the temporal residuals after the deformations, which lowers the learning curve especially at the early training phase. To realize this insight, we proposed the following method inspired from optical flow[2]:

```
value, flow = FourCastNeXt(input)

output = value + temporal_warp(input, flow)
```

Value and flow are computed via separate output heads on top of the common network backbone used by FourCastNet. Value represents the current state of the system and flow represents the change over the current state. We also initialize the flow head to near-zero weights which results in near identity warping at the start of training. In other words, the model initially learns from the pixel-wise difference between the neighbouring temporal steps, relying upon persistence to provide skill, but during training learns its own flow field. Our further analysis in Section 5 shows that this method dramatically speeds up the training convergence and achieves better final error compared to the NVLab Baseline.

### 3.5 Multi-step fine-tuning

The NVLab baseline FourCastNet was trained on single time steps. When predicting multiple time steps into the future, the prediction errors tend to accumulate over time. We propose to fine-tune the model on multiple time steps after the initial training on single time steps. We employ a curriculum learning strategy[3] that sequentially fine-tunes for more time steps. However, we noticed that naively fine-tuning the model for more time steps results in a catastrophic forgetting problem, such that the error of early time steps increases quickly after the fine-tuning. To prevent this catastrophic forgetting problem, we designed a fine-tuning algorithm that incorporates a prior preservation loss (Ruiz et al., 2023) via the following frozen teacher model:

```
Initialize the teacher and student model to the one pre-trained on single time steps.

Freeze the weights of the teacher model

for curr_step = 1 ... max_subsequent_steps
    while student not converged:
        teacher_step = random sample between 1 and curr_step
        loading training example obs1 at t(1), obs2 at t(teacher_step-1)
            and target at t(teacher_step)
        teacher_output = auto-regressive run teacher(obs1) to teacher_step – 1
        multi_step_loss = loss(student(teacher_output), target)
        single_step_loss = loss(student(obs2), target)
        loss = multi_step_loss + single_step_loss
    teacher = student
    freeze the weights of the teacher model
```

---

[2] For a general introduction to optical flow, refer to Niehorster (2021).
[3] For a description and detailed review of curriculum learning strategies, see Soviany et al. (2022).

As can be seen from the above algorithm, the *single_step_loss* enforces a minimisation of loss for single time stepping, and *multi_step_loss* provides a minimisation after error accumulation from multiple model steps. As the number of steps prior to *multi_step_loss* is randomized between loss steps, the model is incentivized to not overfit for that particular lead time. Slowly increasing the maximum value of these steps allows the model to slowly fine tune to deeper auto recurrent steps.

## 4. Experiments and Results

### 4.1 Training details

The variables used in our experiment are the same as those utilized by Pathak et al. (2022b) in the original FourCastNet paper. They comprise 20 variables (listed in Section 4.3.2) representing aspects of the state of the atmosphere, taken from the ERA5 data set. See Pathak et al. (2022b) for a detailed breakdown of the variables and the reasoning for their inclusion.

We train our model in two stages. Stage 1 is to pre-train the model on single time steps for a total of 40,000 steps, using around 160,000 total training examples. Once the first stage is complete, we fine-tune the model for 2,000 steps for each increment of prediction steps until a maximum step of 4, representing a 24-hour period. The total number of training steps for both stages combined is 48,000.

We use the LAMB optimizer (You et al., 2020) with cosine learning rate schedule for all our training stages. The initial learning rate for pre-training is 3e-3 and the final learning rate is 3e-4. The initial learning rate for fine-tuning is 1e-4 and the final learning rate is 1e-5. The total batch size follows a schedule[4] that starts with 4 and doubles from step 12,000. The model is trained with Pytorch Lightning using a Distributed Data Parallel (DDP) strategy with fp16 mixed precision.

### 4.2 Computational requirements

Our training cluster consisted of four NVIDIA V100 SXM2 GPUs (NVIDIA Corporation, 2020) with 32GB memory each. The total training wall time for FourCastNeXt, including both single-step and multi-step, is about 35 hours.

In comparison, the NVLab baseline FourCastNet was trained on a cluster of 64 NVIDIA A100 SXM GPUs (NVIDIA Corporation, 2021) with 80GB memory each. The total training wall time was about 16 hours.

The A100 GPUs are significantly more powerful than the V100, but are 4-5 times more expensive. At the time of writing, many researchers will be unable to access any A100 GPUs, let alone significant numbers of them. This may change in coming years, but at present this represents a major barrier to entry for would-be researchers.

It is difficult to establish a precise, repeatable efficiency factor for comparing the two GPUs. Technical profiling data for the FourCastNet experiment is not available, and regardless such numbers are subject to change due to various confounding factors or if details of the model or training regime are changed. As such, researchers should regard this section as a general guide for establishing the improved training efficiency of FourCastNeXt.

---

[4] For an explanation of why a schedule for selecting batch sizes is effective, see e.g. Chowdhery et al. (2022), Smith et al. (2018) and McCandlish et al. (2018).

Lambda (Balaban, 2021) provides a performance comparison between the V100 GPU and the A100 GPU, including various configurations (e.g. different numbers of GPUs in parallel) on different deep learning experiments. NVIDIA Corporation (2021) also provides benchmarking data on the V100 and A100. None of the benchmarked models directly match FourCastNet or FourCastNeXt. Taking into account these benchmarks, and considering apparent performance in practice, we have chosen to use a factor of 3 to express the general performance difference – that is, we conservatively regard an A100 to be three times faster than a V100 for the work at hand.

FourCastNeXt utilized 140 GPU-hours (4 GPUs x 35 hours). FourCastNet utilized 1024 GPU-hours (64 GPUs x 16 hours), but these are not yet equivalent due to the improved efficiency of the A100 GPU. The ratio for comparison then is 140 compared to 3072 (1024 GPU hours x 3 efficiency factor). This provides a general estimate of 5% of the original computational requirement.

| *Model* | *GPU type* | *Number of GPUs* | *Training wall time* | *Total GPU wall time* | *Normalized to estimated V100 GPU hours* |
|---|---|---|---|---|---|
| FourCastNet | A100 | 64 | About 16 hours | $64 \times 16 = 1024$ hours | $1024 \times 3 = 3072$ hours |
| FourCastNeXt | V100 | 4 | About 35 hours | $4 \times 35 = 140$ hours | 140 hours |

**Table 1:** Summary of GPU configuration and computational capacity of training clusters

## 4.3 Results

We report all our results from the checkpoint of the final training step. Testing data extends from 1 Jan 2018 to 31 Dec 2018, sampled every 6 hours. This matches FourCastNet to allow direct comparison. RMSE is computed for (a) NVLab baseline FourCastNet against ERA5 and (b) our model against ERA5. To compute the testing RMSE, we firstly unstandardize the network predictions by multiplying the standard deviations followed by adding the means. Then, we use *torch.sqrt(F.mse_loss(predictions, ground_truth))* to compute the RMSE. The mean, standard deviation and the pretrained weights of the NVLab baseline FourCastNet model are provided by Pathak et al. (2022a).

Let:

- *NVLab baseline* denote the NVLab FourCastNet model pretrained by NVLab.
- *baseline@5%* denote our training of the FourCastNet model using about 5% compute, achieved by reducing the number of training steps to 40,000 (as described as stage 1 in Section 4.1) and using four V100 SXM2 GPUs, includes the large training set, but excludes any other modifications.
- *deep-norm init* denote our FourCastNeXt model trained for single steps and 5% compute, includes the large training set, includes deep-norm initialization, but excluding any other modifications
- *single-step pretrained* denote our FourCastNeXt model trained for single steps using the large training set, deep-norm initialization, 4x4 patch size, flow field, (but no multi-step fine-tuning), and 5% compute.
- *4-step fine-tuned (best)* denote our FourCastNeXt model, using the large training set, deep-norm initialization, 4x4 patch size, flow field, fine-tuned for 4 steps after the single-step pretraining, and 5% compute. This is currently our best model. For a discussion of why we selected 4-step fine-tuning as the best model see Section 5.4.

### 4.3.1 Training the NVLab baseline model with 5% compute (baseline@5%)

To provide an initial point of comparison, we trained the *NVLab baseline* model with 5% compute by constraining the number of training steps to 40,000 and GPUs utilized. We observed that the predictions of *baseline@5%* diverged from the original *NVLab baseline* around step 19. The following log-plot of the average RMSE across all variables shows the divergence (Figure 2). We have plotted our results using a logarithmic scale as the RMSE of *baseline@5%* grows exponentially.

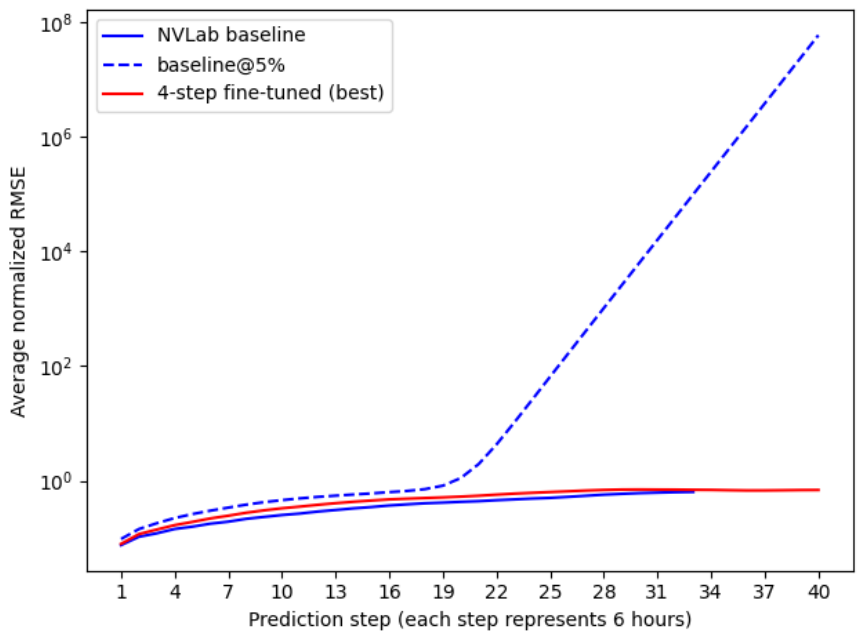

**Figure 2:** *NVLab baseline* vs. *Baseline@5%* vs. *4-step fine-tuned (best)* - as measured by RMSE, normalized and averaged across all modelled ERA5 variables

The *baseline@5%* model suffers an assumed mode collapse around step 19 due to error accumulation and ceases to produce plausible predictions. In contrast, *4-step fine-tuned (best)* closely follows the *NVLab baseline* RMSE and produces reasonable outputs throughout.

We also plot the training curves for *baseline@5%* and *single-step pretrained* (Figure 3).

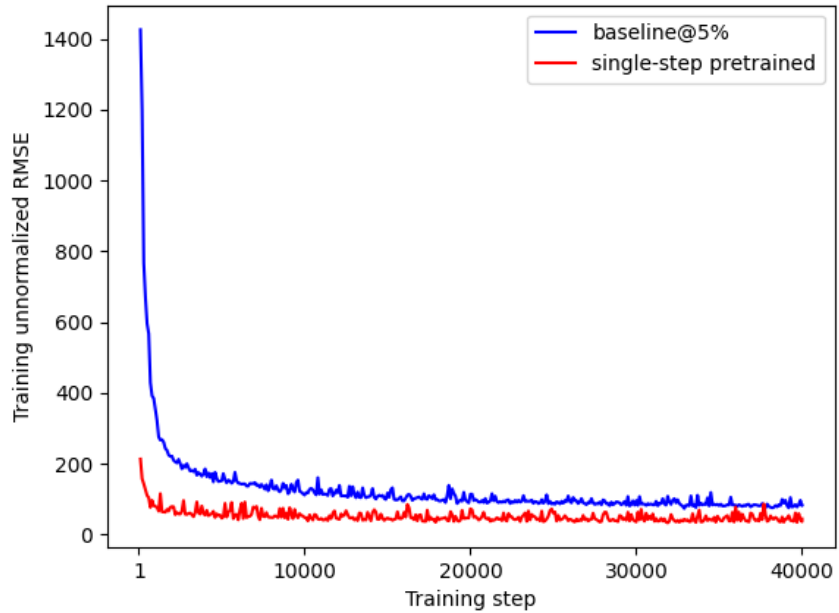

**Figure 3:** Training curves

As can be seen from the above training curve plot, the training RMSE for *single-step pretrained* is consistently lower than *baseline@5%* throughout the entire training process, which demonstrates the effectiveness of our improved methods.

### 4.3.2 Comparisons between NVLab baseline and 4-step fine-tuned (best)

**First prediction step (6 hour lead time) accuracy comparison**

We first present results comparing the first prediction step (6 hours lead time) between *NVlab baseline* and FourCastNeXt with unnormalized RMSE in Table 2. The unnormalized RMSEs allow the comparison of individual variables in their original data ranges. The bold text denotes an improvement over the *NVLab baseline*.

| ERA5 variable | NVLab baseline (unnormalized RMSE) | 4-step fine-tuned (our best) |
|---|---|---|
| 10u (Longitudinal wind speed at 10 metres height) | 0.531046 | 0.564317 |
| 10v (Meridional wind speed at 10 metres height) | 0.526364 | 0.562536 |
| 2t (Atmospheric temperature at 2 metres height) | 0.755716 | 0.759327 |
| sp (Surface pressure) | 137.290480 | **43.658535** |
| msl (Mean sea level pressure) | 38.836962 | 46.771553 |
| t850 (Atmospheric temperature at 850 hPa) | 0.512118 | 0.525746 |
| u1000 (Longitudinal wind speed at 1000 hPa) | 0.582749 | 0.616646 |
| v1000 (Meridional wind speed at 1000 hPa) | 0.578866 | 0.614807 |
| z1000 (Geopotential at 1000hPa) | 29.882532 | 36.547699 |
| u850 (Longitudinal wind speed at 850 hPa) | 0.792813 | 0.820308 |
| v850 (Meridional wind speed at 850 hPa) | 0.775644 | 0.805753 |
| z850 (Geopotential at 850hPa) | 26.972977 | 34.094856 |
| u500 (Longitudinal wind speed at 500 hPa) | 1.070337 | 1.150244 |
| v500 (Meridional wind speed at 500 hPa) | 1.083850 | 1.157692 |
| z500 (Geopotential at 500hPa) | 33.929530 | 43.161873 |
| t500 (Atmospheric temperature at 500 hPa) | 0.399981 | 0.427304 |
| z50 (Geopotential at 50 hPa) | 79.489610 | **78.297844** |
| r500 (Relative humidity at 500 hPa) | 6.551108 | 6.946574 |
| r850 (Relative humidity at 850 hPa) | 5.525181 | 5.582663 |
| tcwv (Total column water vapour) | 0.821443 | 0.825033 |

**Table 2:** First prediction step results (unnormalized, lead time 6 hours). The bold text denotes an improvement over the *NVLab baseline*.

Next, we present results with normalized RMSEs in Table 3 that allow comparison across variables. The bold text denotes an improvement over the *NVLab baseline*.

| *ERA5 variable* | *NVLab baseline (normalized RMSE)* | *4-step fine-tuned (our best)* |
|---|---|---|
| 10u (Longitudinal wind speed at 10 metres height) | 0.094613 | 0.100541 |
| 10v (Meridional wind speed at 10 metres height) | 0.110120 | 0.117688 |
| 2t (Atmospheric temperature at 2 metres height) | 0.035274 | 0.035443 |
| sp (Surface pressure) | 0.014199 | **0.004515** |
| msl (Mean sea level pressure) | 0.028015 | 0.033738 |
| t850 (Atmospheric temperature at 850 hPa) | 0.032535 | 0.033401 |
| u1000 (Longitudinal wind speed at 1000 hPa) | 0.094045 | 0.099515 |
| v1000 (Meridional wind speed at 1000 hPa) | 0.109024 | 0.115793 |
| z1000 (Geopotential at 1000hPa) | 0.026457 | 0.032358 |
| u850 (Longitudinal wind speed at 850 hPa) | 0.095732 | 0.099052 |
| v850 (Meridional wind speed at 850 hPa) | 0.124102 | 0.128919 |
| z850 (Geopotential at 850hPa) | 0.017288 | 0.021852 |
| u500 (Longitudinal wind speed at 500 hPa) | 0.089858 | 0.096566 |
| v500 (Meridional wind speed at 500 hPa) | 0.118515 | 0.126589 |
| z500 (Geopotential at 500hPa) | 0.009864 | 0.012548 |
| t500 (Atmospheric temperature at 500 hPa) | 0.034644 | 0.037011 |
| z50 (Geopotential at 50 hPa) | 0.011859 | **0.011682** |
| r500 (Relative humidity at 500 hPa) | 0.195522 | 0.207325 |
| r850 (Relative humidity at 850 hPa) | 0.209885 | 0.212069 |
| tcwv (Total column water vapour) | 0.050537 | 0.050758 |
| *Average normalized RMSE* | *0.075104* | *0.078868* |

**Table 3**: First prediction step results (normalized, lead time 6 hours). The bold text denotes an improvement over the *NVLab baseline*.

The comparison at the first prediction step (lead time 6 hours) suggests that the performance of our model is mostly on par with the *NVLab baseline* for most of the variables. Notably, variable *sp* and *z50* outperform the *NVLab baseline*. We acknowledge an anomalously high level of accuracy in surface pressure compared to the baseline. An analysis of this is left to future work.

**Longer lead time (multiple prediction steps) comparison**

The following plot (Figure 4) shows the average normalized RMSE over all predicted ERA5 variables versus the prediction steps.

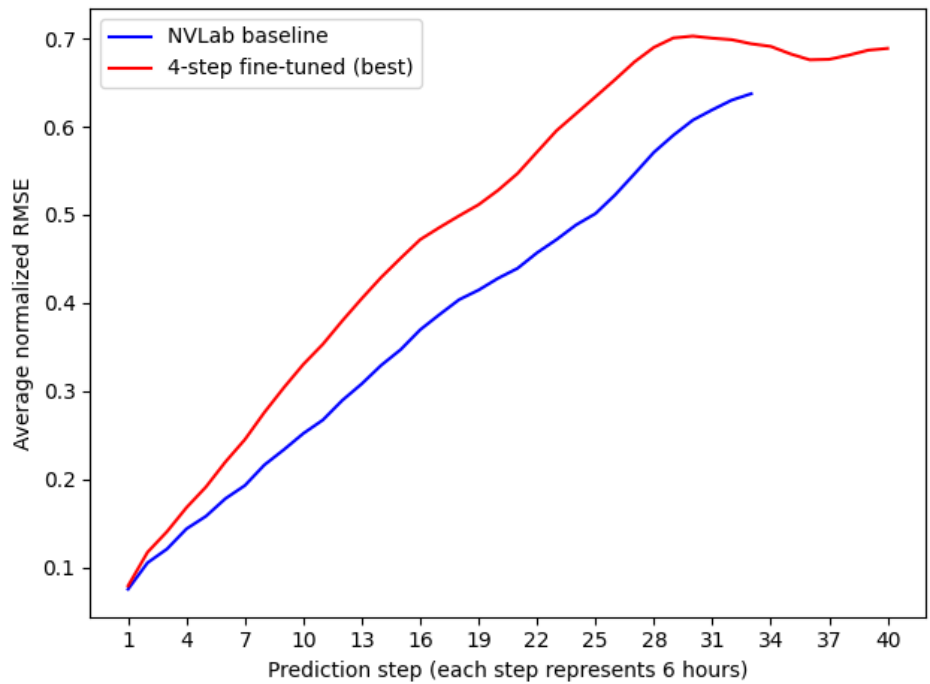

**Figure 4:** *4-step fine-tuned (best)* vs. *NVLab baseline* – as measured by RMSE, normalized and averaged across all modelled ERA5 variables

As can be seen from Figure 4, the *4-step fine-tuned (best)* model provides reasonable results out to prediction step 40 (240 hours). The original NVLab paper only considered periods out to around prediction step 33 (198 hours) so we concluded our analysis of the *NVLab baseline* at that time step for consistency. We also examined the RMSEs for a few individual variables and they are consistent with this pattern. Figures 5 and 6 provide two examples: variables *2t* and *tcwv*. Other variables followed similar patterns.

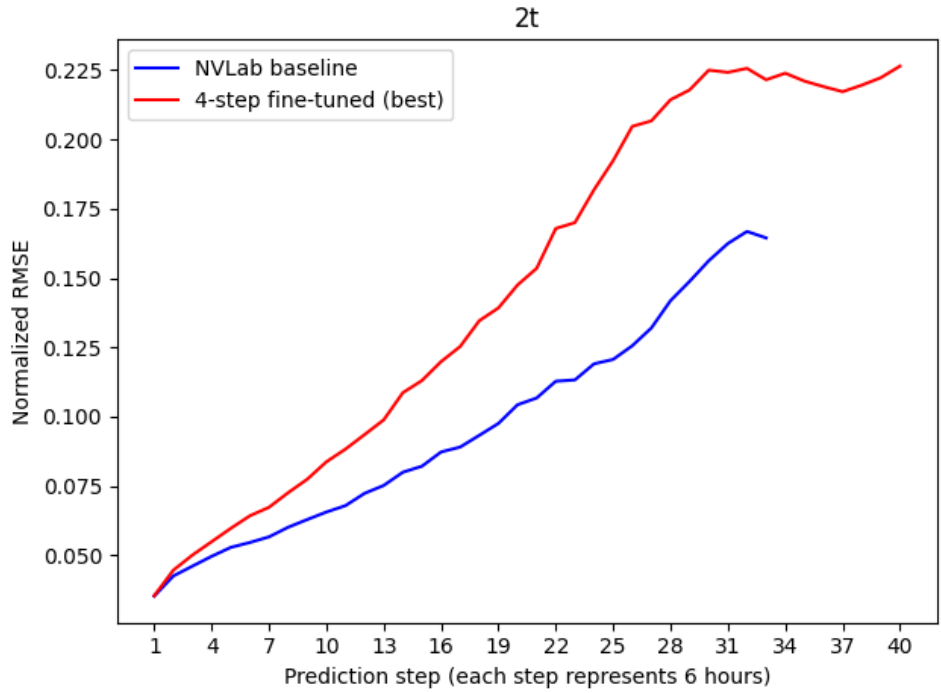

**Figure 5:** *4-step fine-tuned (best)* vs. *NVLab baseline* at different prediction steps, showing normalized RMSE for variable *2t*

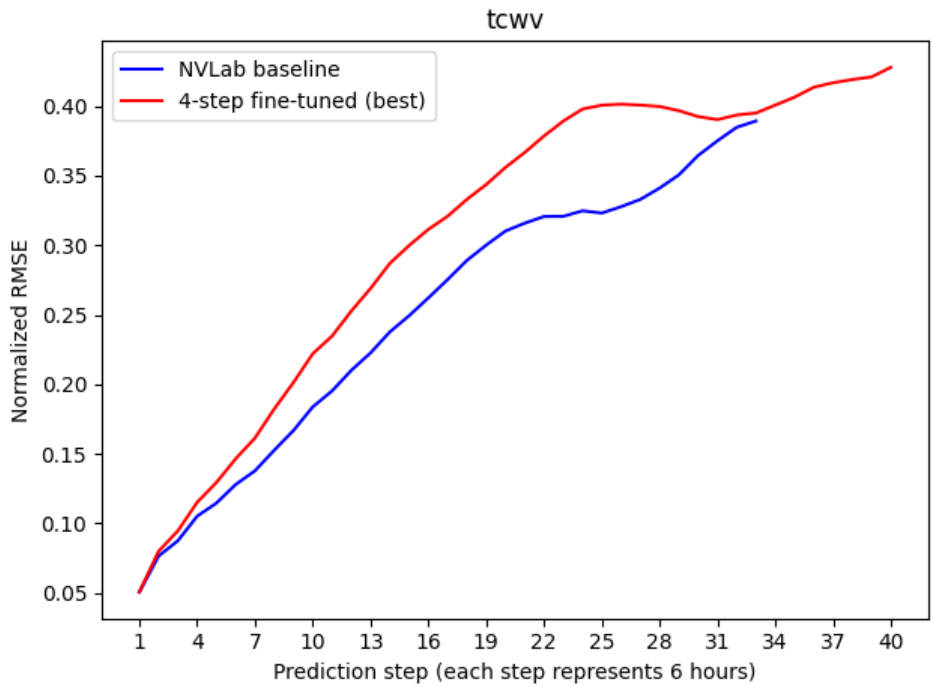

**Figure 6:** *4-step fine-tuned (best)* vs. *NVLab baseline* at different prediction steps, showing normalized RMSE for variable *tcwv*

# 5. Contribution of Methods

In this section, we investigate the contribution of the methods discussed in Section 3. For the analysis in Sections 5.1 and 5.2, 1500 training steps were undertaken.

## 5.1 Deep-norm initialization versus baseline

The following plot (Figure 7) is the training loss of deep norm initialization versus the baseline.

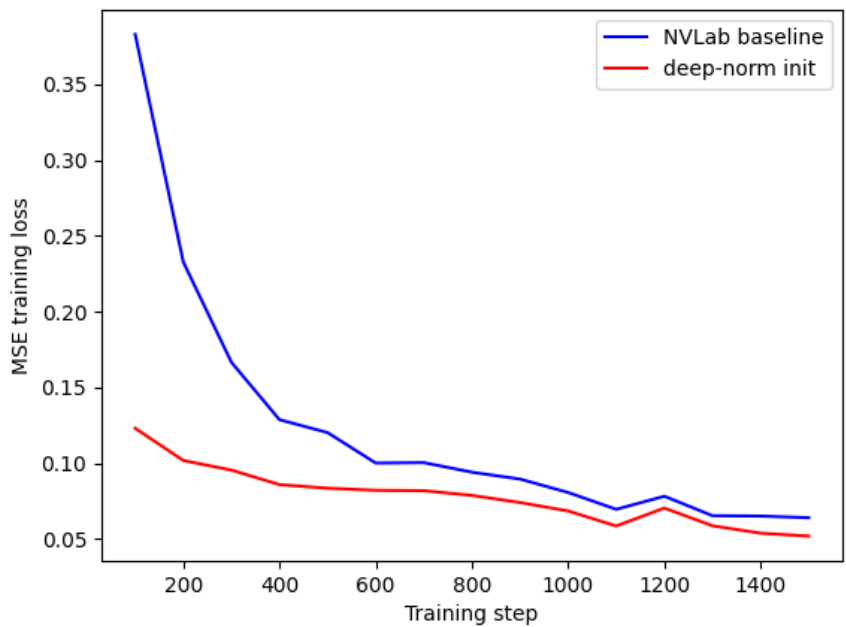

**Figure 7:** Deep-norm initialization vs. *NVLab baseline*, according to mean squared error (MSE) training loss.

As can be seen from the above plot (Figure 7), deep-norm initialization stabilizes the training at the early phase and speeds up the training process by reaching the same level of training loss faster than the baseline and therefore allows training to be stopped after fewer training steps.

## 5.2 Patch size 4x4 and flow field

In this section, we show that a smaller patch size of 4x4 and flow field further improve convergence during training in addition to the deep-norm initialization.

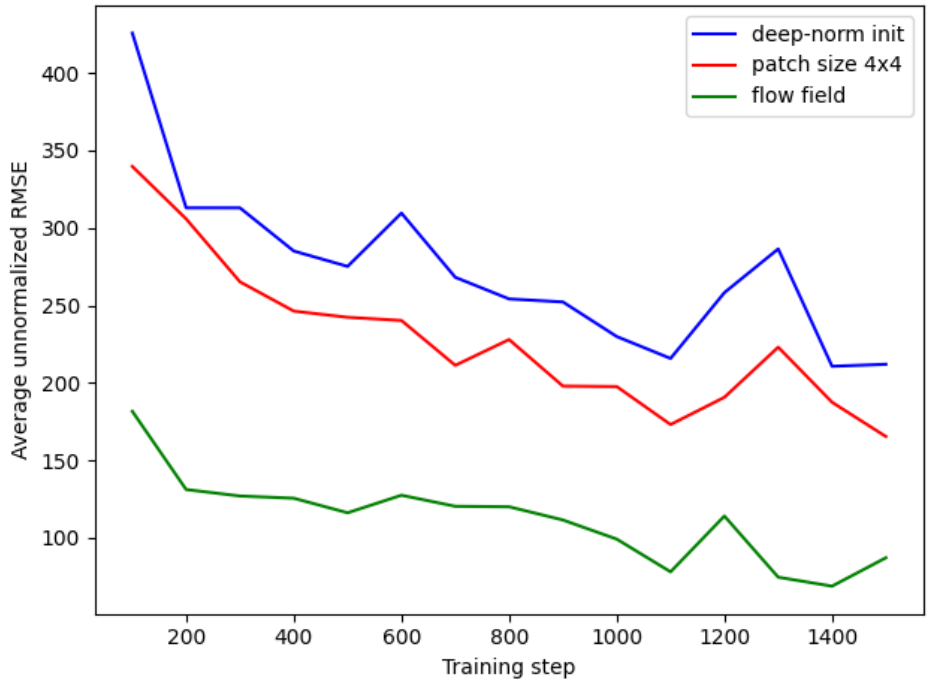

**Figure 8:** Contributions of each technique as it is combined into the model. Here, *deep-norm init* refers to the model with deep-norm initialization and a patch size of 8x8. *Patch size 4x4* refers to the model with both deep-norm initialization and a patch size of 4x4. *Flow field* refers to the model with deep-norm initialization, a patch size of 4x4, and flow field.

As can be seen from the above plot (Figure 8), reducing the patch size and adding flow field consistently outperforms deep-norm initialization alone. In addition, flow field yields the largest reduction in RMSE.

Discussion of the effect of the 4x4 patch size on fine-scale structure is included in Section 6.

## 5.3 Four-step fine-tuning versus single-step pretraining

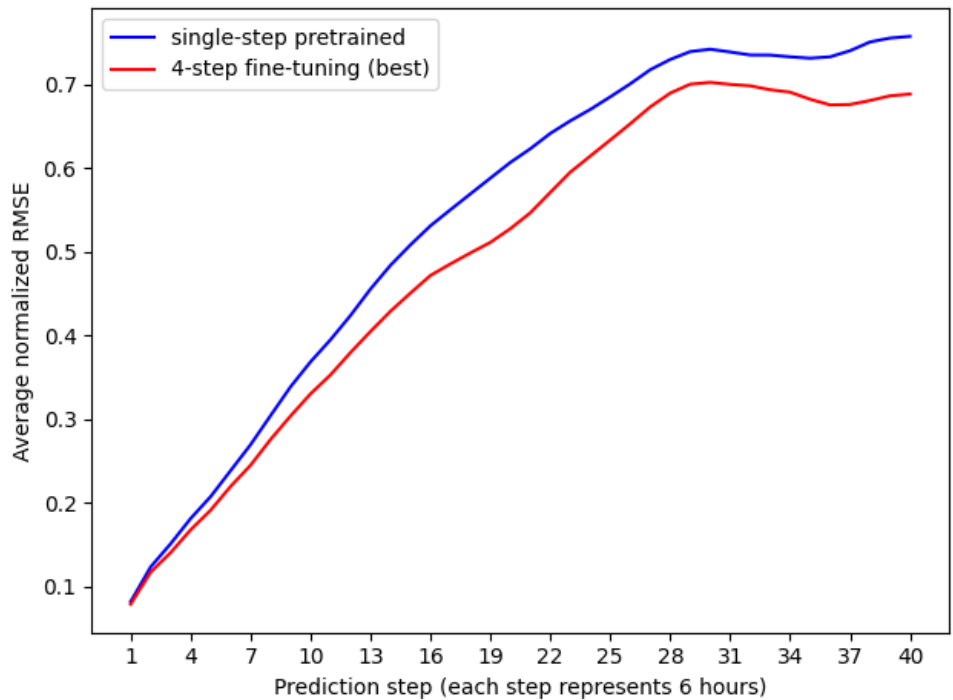

**Figure 9:** *Four-step fine-tuned (best)* vs. *single-step pretrained* - as measured by RMSE, normalized and averaged across all modelled ERA5 variables

As can be seen from the above plot (Figure 9), four-step fine-tuning consistently improves the average normalized RMSE for predictions over a long horizon.

## 5.4 Fine-tuning for different number of steps

Experiments initially suggested that 4-step fine-tuning would be most suitable. We then attempted to fine-tune the model at various numbers of fine-tuning steps, using the fine-tuning technique described in Section 3.5. Plots of average normalized and unnormalized RMSE versus prediction steps for different degrees of fine-tuning are as follows (Figures 10 and 11):

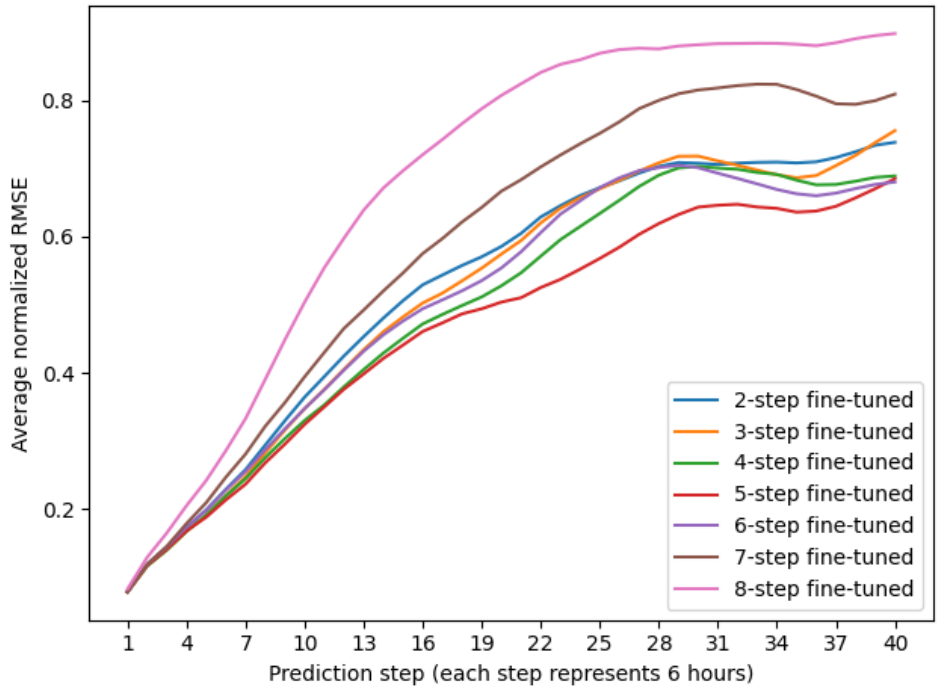

**Figure 10:** Comparing different degrees of fine-tuning, showing accuracy over prediction steps (lead time) as measured by average normalized RMSE.

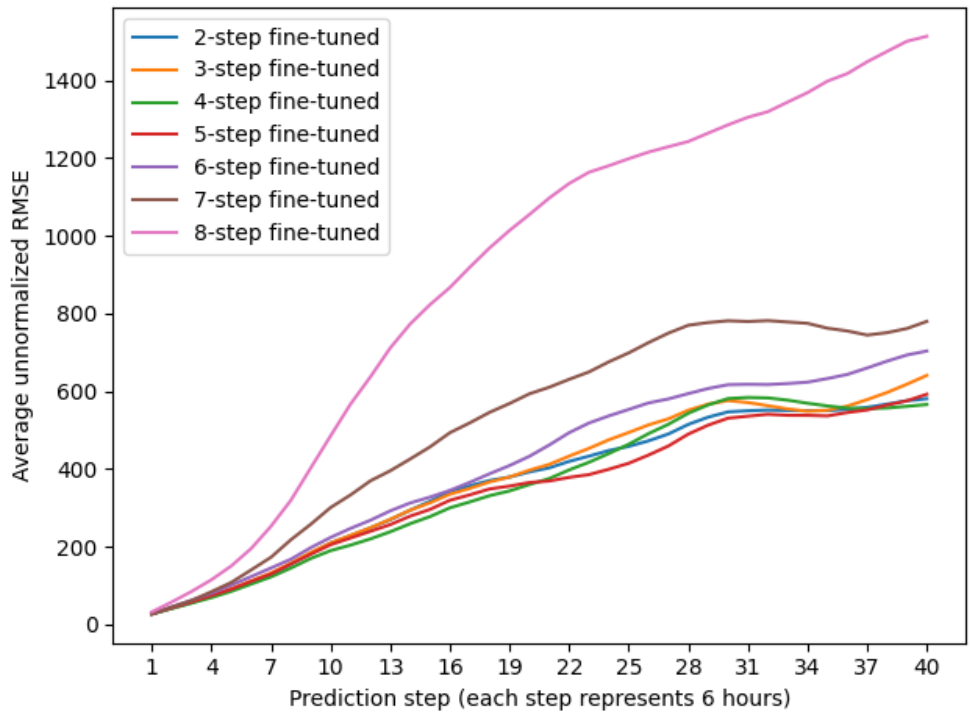

**Figure 11:** Comparing different degrees of fine-tuning, showing accuracy over prediction steps (lead time) as measured by average unnormalized RMSE. The differences apparent here compared to Figure 10 shows that optimal performance is different depending on the variable considered.

Table 4 below provides a numerical comparison between 4-step fine-tuning and 5-step fine-tuning from prediction steps 1 to 5.

| Prediction step | 4-step average normalized RMSE | 5-step average normalized RMSE |
|---|---|---|
| 1 (6 hours) | **0.078868** | 0.078990 |
| 2 (12 hours) | **0.117048** | 0.117789 |
| 3 (18 hours) | **0.140460** | 0.141704 |
| 4 (24 hours) | **0.167720** | 0.168399 |
| 5 (30 hours) | 0.191170 | **0.188570** |

**Table 4:** Comparison of 4 and 5 step fine-tuning at early prediction steps. Bold text denotes better performance.

Utilizing more than five steps of fine-tuning does not yield better results. We attribute this phenomenon to the possibility that the model architecture lacks the ability to capture long-range temporal dependencies.

Comparing between four and five fine-tuning steps shows mixed results, and normalizing the RMSE values is significant in this case. As such, the preference between four-step and five-step fine-tuning will depend on the use case, the lead time of interest, and which variables are of greatest interest. Four steps of fine-tuning provides reasonable results and a more detailed comparison of four and five step fine-tuning could be considered in future work.

## 6. Physical Realism

Both FourCastNeXt and FourCastNet provide statistically credible models for estimating the state of the atmosphere, for the variables they cover, at lead times out to 198 hours, at 0.25 degrees resolution. FourCastNeXt was trained to provide useful results out to 240 hours. Neither FourCastNeXt nor FourCastNet provide best-in-class predictions compared to more recent NESM model innovations, but most of those alternatives are even more computationally expensive to train. As such, FourCastNeXt continues to fill a valuable niche in producing reasonable results for a very low comparative training cost.

As the focus of this paper is on the methods and effectiveness of the FourCastNeXt optimization techniques, a full assessment of the physical realism of FourCastNeXt outputs is beyond scope. However, this section briefly discusses some of the strengths and limitations we noted in FourCastNeXt and FourCastNet.

The strengths and weaknesses discussed in this section also serve as examples of some of the types of issues that researchers should be aware of when exploring NESMs[5]. NESMs are still at an early stage of research and should be used with care despite their impressive capabilities.

### 6.1 Behaviour in the southern polar region

FourCastNeXt shows better behaviour in the southern polar region compared to FourCastNet (see Figure 12). The physical grid spacing used by many earth system models (including ERA5) around the pole approaches zero, making it mathematically challenging. Some other NESMs have approached this issue by moving away from training on a regular grid, some to explicit graphs (e.g. Lam et al., 2023) and others to a spherical coordinate system (Bonev, Kurth, et al., 2023), including FourCastNet V2 (Bonev, Hundt, et al., 2023).

---

[5] For a discussion of some factors that can potentially impact the robustness and reliability of machine learning applied to weather and climate modelling, see Section 10 of de Burgh-Day and Leeuwenburg (2023).

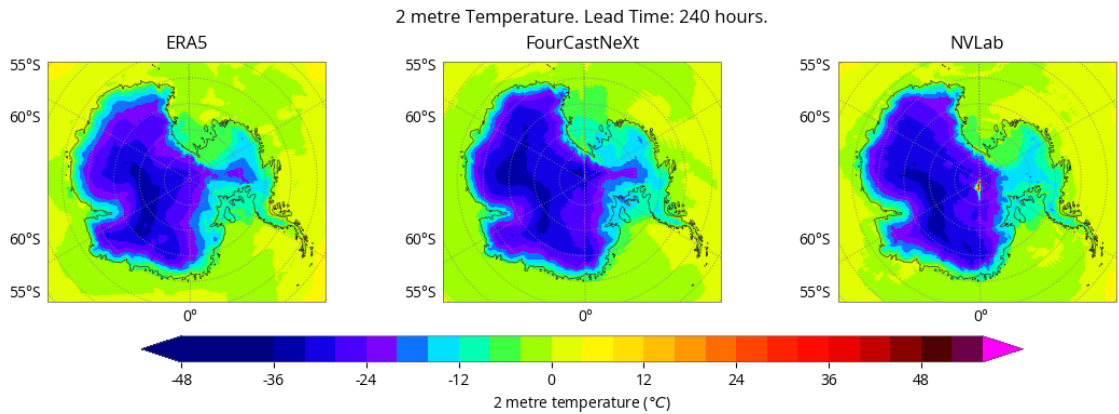

**Figure 12:** This shows the southern polar region at a lead time of 10 days (240 hours, 40 prediction steps), showing the appearance of artefacts in the NVLab baseline, which are not present in the FourCastNeXt model. Created using `earthkit.maps` (ECMWF, 2023)

## 6.2 Illustration of lead time

FourCastNeXt shows good results out to 240 hours (10 days, 40 prediction steps), which is longer than FourCastNet was explicitly trained to provide. See Figures 2 and 13 for relevant illustrations of performance over the progression of the model over its lead time.

FourCastNeXt: Optimizing FourCastNet Training for Limited Compute    https://doi.org/10.48550/arXiv.2401.05584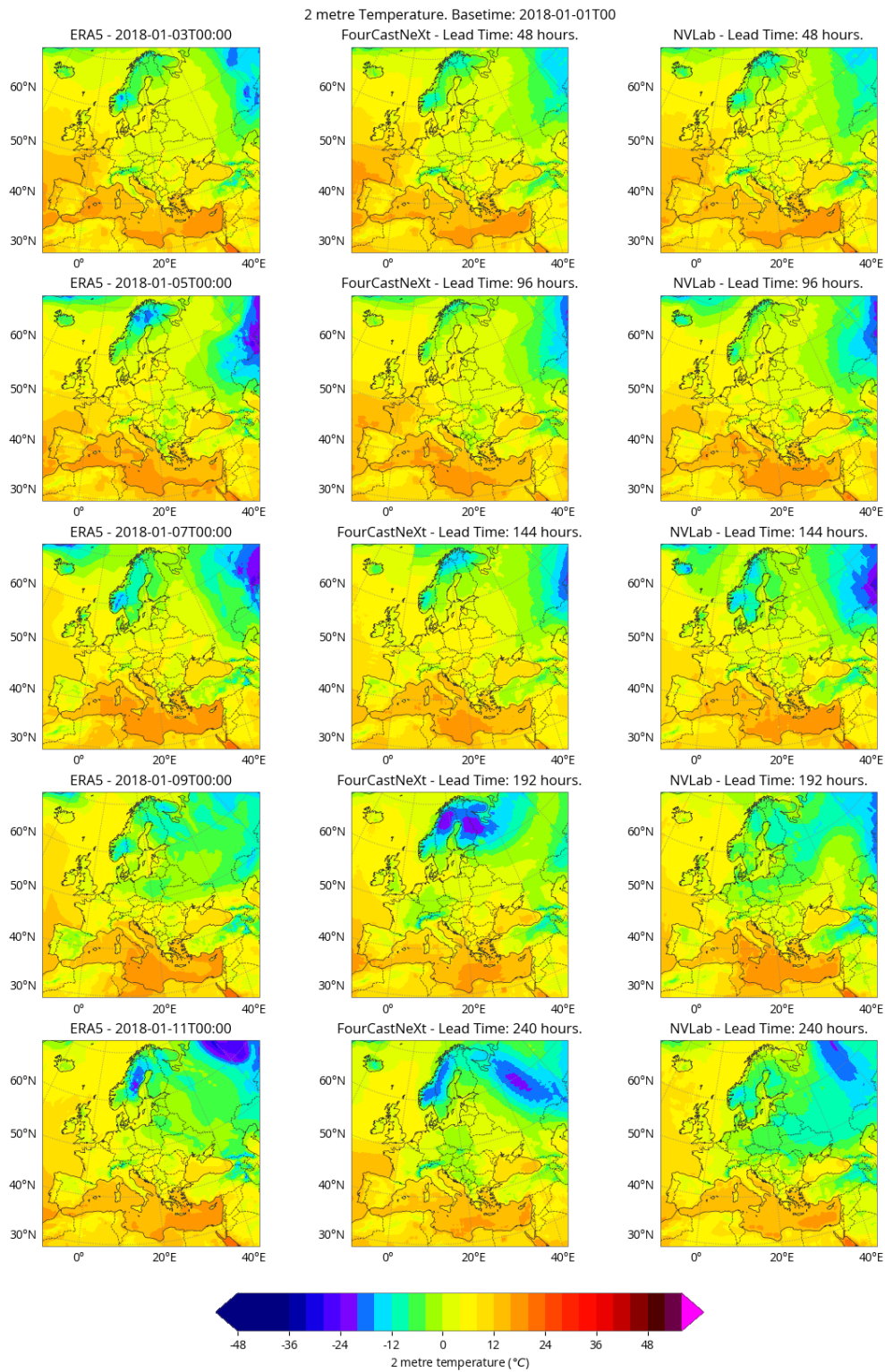

**Figure 13:** Shows a case study over Europe to a long lead time (10 days, 240 hours, 40 prediction steps) where FourCastNeXt comparably captures the weather situation to FourCastNet. This is a selected example only and shouldn't be assumed to be representative of general performance. Created using `earthkit.maps` (ECMWF, 2023)

## 6.3 Discontinuities at patch boundaries

FourCastNeXt shows discontinuities at the patch boundaries when stitching together the subset crops of 640 x 1280. This is particularly apparent around 180 degrees longitude. See Figure 14 for a side-by-side comparison between FourCastNeXt and the NVLab baseline. It is thought that with some attention to how patches are stitched together at inference time - either through boundary merging or interpolation - this issue may be resolved.

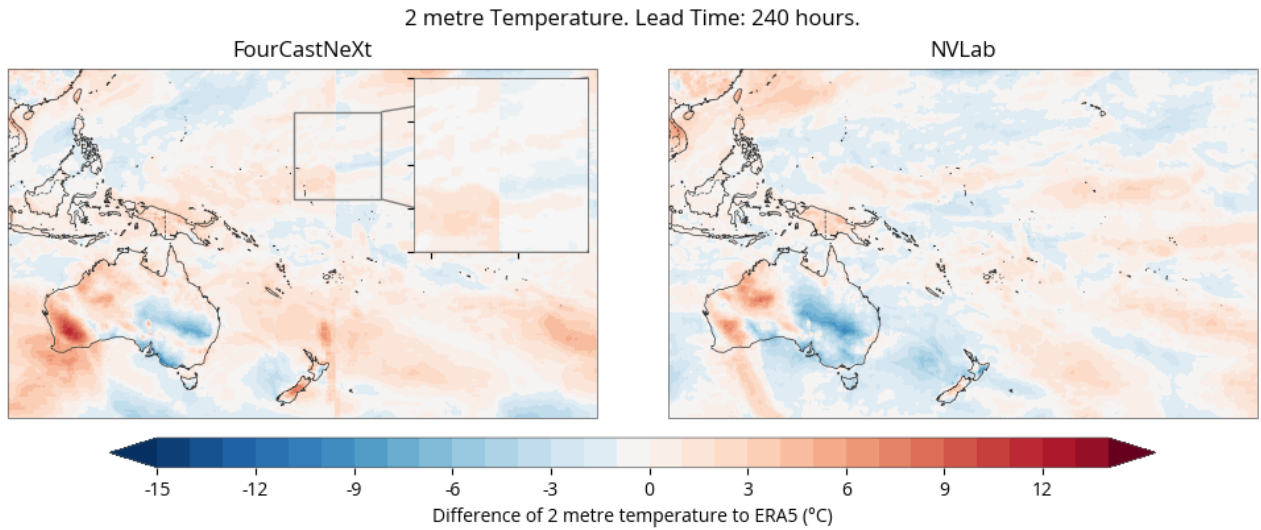

**Figure 14:** Shows the temperature discontinuity which appears around 180 degrees longitude. This is likely to be a side-effect of patching random 640 x 1280 crops back together after inference. It is also illustrative of a type of artefact which is unlikely to occur in a physical NWP model, and highlights the importance of fine-grained evaluation of machine learning outputs which go beyond global aggregate statistics.

## 6.4 Fine-scale image structure

We earlier speculated that using a 4x4 patch size would improve the fine-scale image structure, compared to larger patch sizes. Plots of outputs do not show a marked qualitative difference in fine-scale structure compared to the NVLab baseline, implying comparable overall quality. Figure 13 provides a visualisation of FourCastNeXt output at various lead times, and Figure 14 provides a side-by-side comparison between FourCastNeXt and FourCastNet. However, as presented earlier, global RMSE statistics are improved compared to the baseline@5%.

As such, it will be necessary to conduct an analysis of the structures at various spatial scales using other approaches. A power spectra analysis (spatial) (see Figures 15a and 15b) is one such approach. Figure 15b shows excellent agreement between FourCastNeXt, FourCastNet and ERA5 for the first 20 harmonics. Figure 15a shows comparable power at the finer spatial scales (harmonic 250 and above) between FourCastNeXt and FourCastNet, noting that both fall below ERA5 (the target). Further work to better quantify and discriminate fine-scale structure as a component of error could be considered. Figures 15a and 15b examine just one variable, at a single lead time, from a single forecast run. While figures 15a and 15b are illustrative, a more complete analysis could compare more variables and more examples to get a better sense of performance overall.

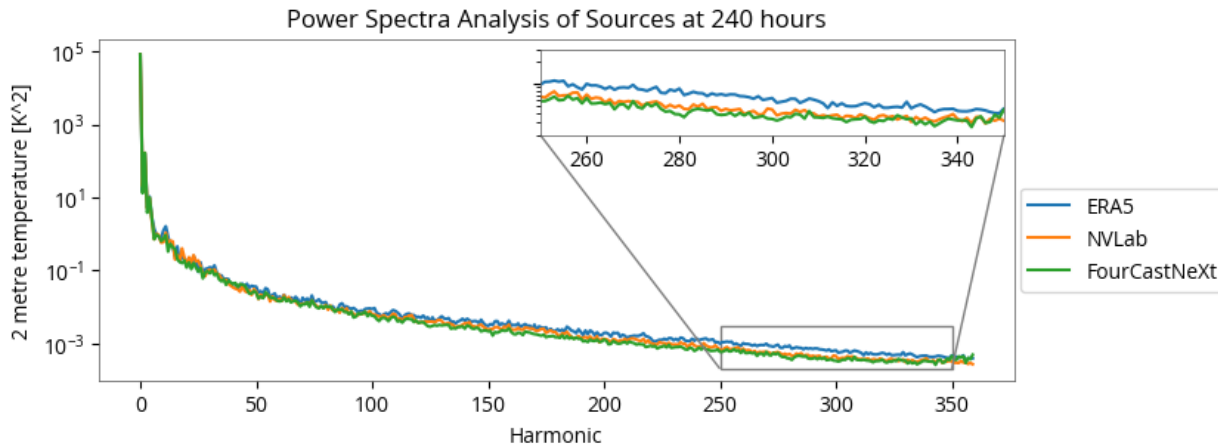

**Figure 15a:** A spherical harmonic power spectra analysis of *2t* at a lead time of 240 hours. Smaller-degree harmonics show the density of larger physical scale features, while higher-degree harmonics show the density of finer scale features. While neither model matches the density of ERA5 at harmonics 250 and above, FourCastNeXt is comparable to the NVLab baseline.

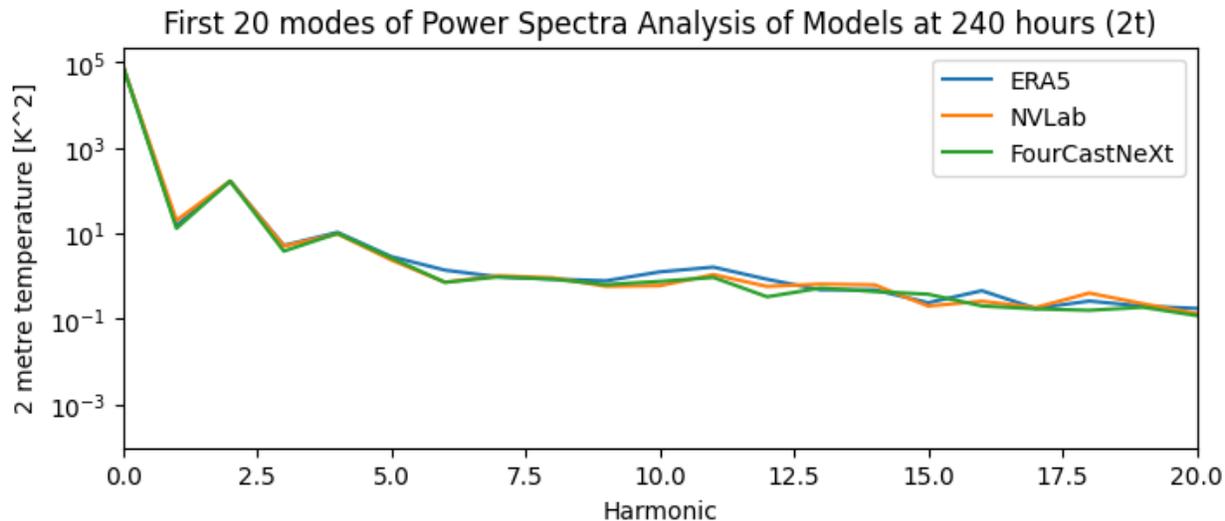

**Figure 15b:** Detail of the first 20 modes, from Figure 15a. Both NVLab and FourCastNeXt show good agreement with ERA5 at larger spatial scales.

In summary, while the image discontinuity artefact (see Figure 14) discussed earlier may be directly attributable to one of the optimization techniques (particularly the disconnect between neighbouring patches of reduced size), overall, the optimizations explored have nonetheless allowed FourCastNeXt to train with about 5% of the compute and reach a comparable level of performance to FourCastNet.

## 7. Conclusions and Future Work

This work investigated a collection of methods to improve the efficiency of training the NVLab baseline FourCastNet. Our model, FourCastNeXt, achieved a comparable level of performance (as measured by RMSE) to the NVLab baseline FourCastNet, while achieving this result using greatly reduced compute resources.

The authors are not aware of any other model at 0.25 degrees resolution which provides usefully accurate predictions with such a comparatively low training cost. The authors hope that FourCastNeXt will provide a more accessible model for a wider range of researchers, and potentially encourage NESM researchers to continue to develop low-cost models to support the research community. Low-cost models should provide a pathway to test and evaluate new hypotheses, experiment with integrating new model inputs, or integrate new data sources without needing to contend with prohibitive training costs.

We hypothesize that the techniques employed in this work may be applicable to other model architectures in the weather and climate domain, and exploration of this could be undertaken.

In addition, the community would benefit from scientific evaluation of FourCastNeXt on case studies and from a physical perspective.

## 8. Software and Data Availability

The scripts used to produce all the results and diagrams for FourCastNeXt (our model) are available from NCI's FourcastNeXt GitHub repository (NCI 2023d). The exact inputs and experimental results, normalization statistics, and the pre-trained weights used in this report are publicly available through the NCI Data Catalogue (Guo et al., 2023).

All the code and data related to the project, including the ERA5 replicated data set (NCI Australia, 2020), predictions and checkpoints (NCI Australia, 2023a and 2023c), and model and scripts (NCI Australia, 2023d) are freely available.

The original FourCastNet (baseline model) is available from Pathak et al. (2022a).